\documentclass[sn-basic]{sn-jnl}

\jyear{2021}%

\theoremstyle{thmstyleone}%
\theoremstyle{thmstyletwo}%

\theoremstyle{thmstylethree}%

\usepackage{natbib}
\usepackage{booktabs}
\usepackage{soul}
\usepackage{amsmath}
\usepackage{amssymb}
\usepackage{makecell}
\usepackage{bbding}
\usepackage{pifont}
\usepackage{bigstrut}
\usepackage{color}
\usepackage{algorithm}
\usepackage{algpseudocode}
\usepackage{multirow}

\begin{document}

\title[SubFace: Learning with Softmax Approximation for Face Recognition]{SubFace: Learning with Softmax Approximation for Face Recognition}


\author{\fnm{Hongwei} \sur{Xu}}\email{2007xuhongwei@163.com}
\equalcont{These authors contributed equally to this work.}

\author{\fnm{Suncheng} \sur{Xiang}}\email{xiangsuncheng17@sjtu.edu.cn}
\equalcont{These authors contributed equally to this work.}

\author*{\fnm{Dahong} \sur{Qian}}\email{dahong.qian@sjtu.edu.cn}

\affil[]{\orgdiv{School of Biomedical Engineering}, \orgname{Shanghai Jiao Tong University}, \orgaddress{\city{Shanghai}, \postcode{200240}, \country{China}}}



\abstract{The softmax-based loss functions and its variants (\textit{e.g.}, cosface, sphereface, and arcface) significantly
improve the face recognition performance in wild unconstrained scenes. A common practice of these algorithms
is to perform optimizations on the multiplication between the embedding features and the linear transformation
 matrix. However in most cases, the dimension of embedding features is given based on traditional design experience, and there is less-studied on improving performance using the feature itself when giving a fixed size. To address this challenge, this paper presents a softmax approximation method called SubFace, which employs the subspace feature to promote the performance of face recognition.
 Specifically, we dynamically select the non-overlapping subspace
features in each batch during training, and then use the subspace features to approximate full-feature among softmax-based loss, so the discriminability of the deep model can be significantly enhanced for face recognition.
Comprehensive experiments conducted on benchmark datasets demonstrate that our method can
significantly improve the performance of vanilla CNN baseline, which strongly proves the effectiveness of subspace strategy with the margin-based loss\footnote{We will release the code publicly on GitHub after publication.}. }




\maketitle

\section{Introduction}\label{sec1}

{T}{he} introduction of Convolutional Neural Networks (CNNs) has greatly improved the performances on face vision tasks in the past 20 years,
such as data cleaning \citep{45}, model parallel acceleration \citep{46}, face recognition \citep{10,11,15,48}, \textit{etc.} In essence,
designing effective loss functions for the optimization of CNNs is pivotal for such improvements, which has attracted great interests and attention from both academia and industry.

In the field of face recognition, many methods are proposed to find a discriminative embedding space for more robust learning. For example, \cite{45} introduce k sub-centers for each class into the ArcFace loss for noise data clean, which judges whether the current sample is noise face according to the distance between the sample of dominant sub-class and non dominant sub-classes. \cite{46} propose a Partial FC strategy where the negative samples in loss function is replaced by the subset of negative samples for ace recognition training. In addition, \cite{48} propose the soft-triple loss with multiple
centers for each category, which can effectively capture the hidden distribution of data through this specially designed mechanism. More recently, the margin-based methods, \textit{e.g.,} ArcFace~\citep{11} and CosFace~\citep{10}, can also achieve better performance by introducing margin penalty to the loss function, which marks a new milestone of metric learning in the face recognition community.

\begin{figure}[!t]
  \centering
  \includegraphics[width=9cm]{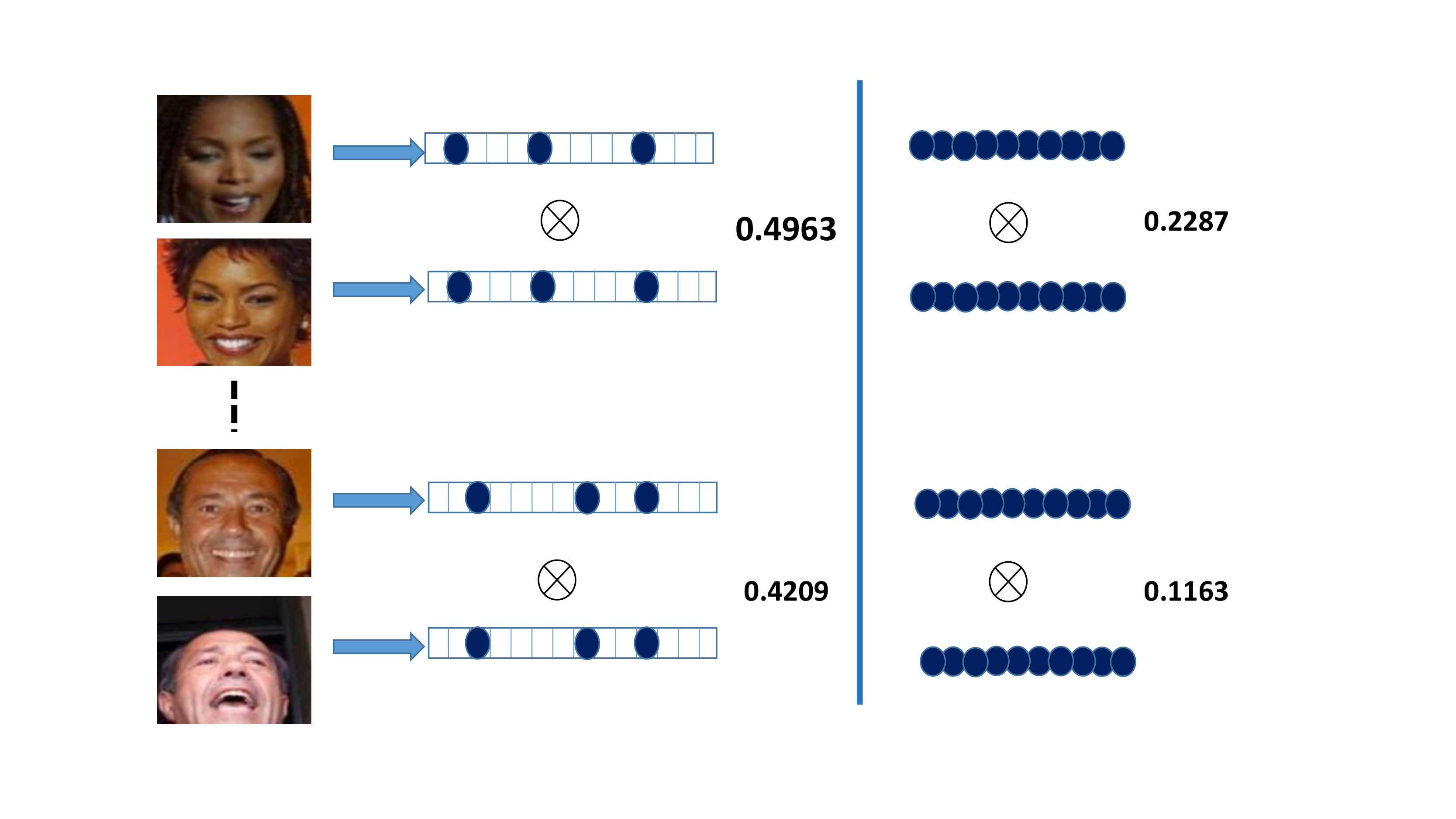}
  \caption{The cos distance of embedding features of the positive image pairs, the dimension is 512, left of the solid line is the cos distance of the 512-d feature, right of the solid line is the min cos distance of the 128-d feature, which is randomly sampled from the 512-d feature.}
  \label{fig1}
\end{figure}

In essence, these methods mainly lay emphasis on employing new feature constraints or designing a better classification interface to achieve competitive performance. To be more specific, they are combined with the softmax loss function to apply to the face recognition scene, which can significantly promote the performance of face recognition in some degree. However, by analyzing the training process of these approaches, we notice that the existing training strategies regard face features as a whole, which fails to pursue the optimization of softmax-based loss on the whole face dataset.
Considering the fact that the face features can be regarded as a high-dimensional vectors, which can help to make the features of sample identities more aggregated. Unfortunately, in some cases we observe that local features are less aggregated than global features, as shown in Fig.~\ref{fig1}. Intuitively, these variations may be caused by the differences in terms of body posture, age, image quality, \textit{etc.}
While the more important reason we thought is that the existing training strategies always ignore the local distinguish ability of face features during training, which may lead to non-uniform distribution of feature discrimination.

To address this challenge, in this paper, we propose a softmax approximation method named SubFace for more robust feature learning.
Firstly, we determine the dimension of subspace, and then dynamically select a subspace of such dimension in the embedding feature and the
corresponding linear transformation matrix. Finally, the subspace is randomly selected at every batch to improve the generalization ability
of sub-feature representation. Extensive experiments demonstrate that our training strategy combined with margin-based loss is very effective and can achieve comparable
results with the state-of-the-arts.

To this end, the major contributions of our work can be summarized as follows:
\begin{itemize}
\item[$\bullet$]We propose a softmax approximation training method named SubFace for more robust feature learning, which can be integrated with the margin-based loss seamlessly for the face recognition.

\item[$\bullet$] On the basis of feature approximation strategy, the subfeature normalization mechanism is introduced to further enhance the discriminability of
learned feature for intra-class and inter-class relations.

\item[$\bullet$]Comprehensive experiments conducted on benchmarks demonstrate that our method can achieve competitive performance when comparing with the state-of-the-arts on face recognition tasks.
\end{itemize}
The remainder of this paper is structured as follows. In Section~\ref{sec2}, we give the related works based on metric-distance-based loss and angular-margin-based loss, and then briefly introduce our method.
 In Section~\ref{sec3}, The details of feature mining and approximation strategy are presented. Extensive evaluations compared with state-of-the-art methods and comprehensive
analyses of the proposed approach are reported in Section~\ref{sec4}. Finally, the conclusion of this paper and discussion of future works are presented in Section~\ref{sec5}.

 \section{Related work}
 \label{sec2}
 From the training process of face recognition methods, the loss functions can greatly promote the performance of face recognition. These methods can be categorized into metric-based loss methods~\citep{15,18,19,20,21,40} and margin-based loss methods~\citep{10,11,12,13,42}.

 \subsection{Metric-based methods}
 Early works usually use the metric-based methods. The common goal of these methods is to learn a distance metric that distinguishes the intra-class and inter-class. These common methods include the contrastive loss and triple losses.
 The contrastive losses~\citep{18,19,20} are those that use paired samples (positive or negative) to train a network to predict whether they belong to the same class. And the triplet loss~\citep{14,39,40,xiang2022learning} consider that the distance between the query and positive samples is greater than that between the query and negative samples by a given distance margin.
 Since contrastive and triplet loss often lead to slow convergence, N-pair loss~\citep{32} is designed, which improves the training convergence by considering the distance between the query samples and other multi negative samples jointly at each update. But these sample-to-samples comparisons often suffer
 from the explosion of sample space. Thus, some sample-to-class schemes, \textit{e.g.}, center loss~\citep{15} and variants~\citep{21}, are proposed to solve the problem of explosive growth in computation, they learn a center for each category and require the same category features to be closed to its center. But these kind methods fail to address the problem of open set in face recognition.

\subsection{Margin-based methods}
Compared to the deep metric learning methods, the angular-margin-based loss~\citep{13,11,42} often remodel the last fully connected layer of classification network to different form. It explicitly adds discriminative constraints to the target logit and makes the same category feature space more compact, it is more efficient and stable,
 L-softmax~\citep{42} constructed a large-margin softmax loss and employed a piecewise function to guarantee the monotonicity of the cosine function. A-softmax~\citep{13} normalized the weight matrix $w$, and learned face features on a hypersphere manifold. In fact,
 L-softmax and A-softmax are difficult to train due to the complex design of the angular margin. To relieve this dilemma, some researchers~\citep{10,11} try to simplify this question. They set the bias term to zero and normalize the weight and embedding feature. The representation of the last fully connected layer is reduced to vector inner product form.
 And Cosface~\citep{10} introduced an additive cosine margin. Arcface~\citep{11} introduced an additive angular margin. More recently, Adaptiveface~\citep{37} and Fariloss~\citep{38} introduce adaptive margin strategy by adjusting the strength of the supervision process during training to address the problem of unbalanced data.

 Compared with existing methods, our proposed SubFace strategy is different from them in the following aspect: (1) Our proposed SubFeature Strategy tries to improve the overall performance of features through mining the discriminative power of local features, supporting it to be more discriminative, while previous researches (such as Arcface and Cosface) mainly lay emphasis on global feature to learn a discriminative model.
 (2) The proposed framework has much fewer parameters and can be embedded into existing face recognition model seamlessly, which make it
more flexible and adaptable in real-world scenarios. (3) To the best of our knowledge, we are among the first attempt to introduce feature approximation mechanism on face recognition task.
 %

\section{Proposed Approach}
\label{sec3}
\subsection{\textbf{Preliminary}}
Usually a training system consists of three parts: training data, feature extraction network, and loss function. And a face training system
can be expressed as $\{\{x_{i},y_{i}\},f,F_{loss}\}$ that contains $N$ number identities, each training sample $x_{i}$ corresponds to a label $y_{i}$,
$f(\cdot )$ is the feature extractor, $f(x_{i}) \in R^d$ represents the embedded feature of i-th face sample.
$F_{loss}$ denotes the loss function during training, which can be represented as:
\begin{align}
  F_{loss}(x_{i},x_{j}) = dis(f(x_{i}),f(x_{j}))
\end{align}

The loss function provides a distance measurement, under which similar targets are as close and non similar targets are far away.
Intuitively, human face has obvious nonlinear characteristics, adopting new loss functions in deep
neural networks is effective approach to make features more discriminative.
However, in addition to the non-linear discriminative embedding, there is clearly locally distinguishable embedding for face recognition. For example, we can distinguish and recognize pedestrians by discriminative parts with attention mechanism~\citep{xiang2020multi}. Inspired by this,
we take a big step forward to explore the identification ability of local features during the training process, which can significantly boost the performance in a large degree for face recognition task.
\begin{figure}[!t]
  \centering
  \includegraphics[width=12cm]{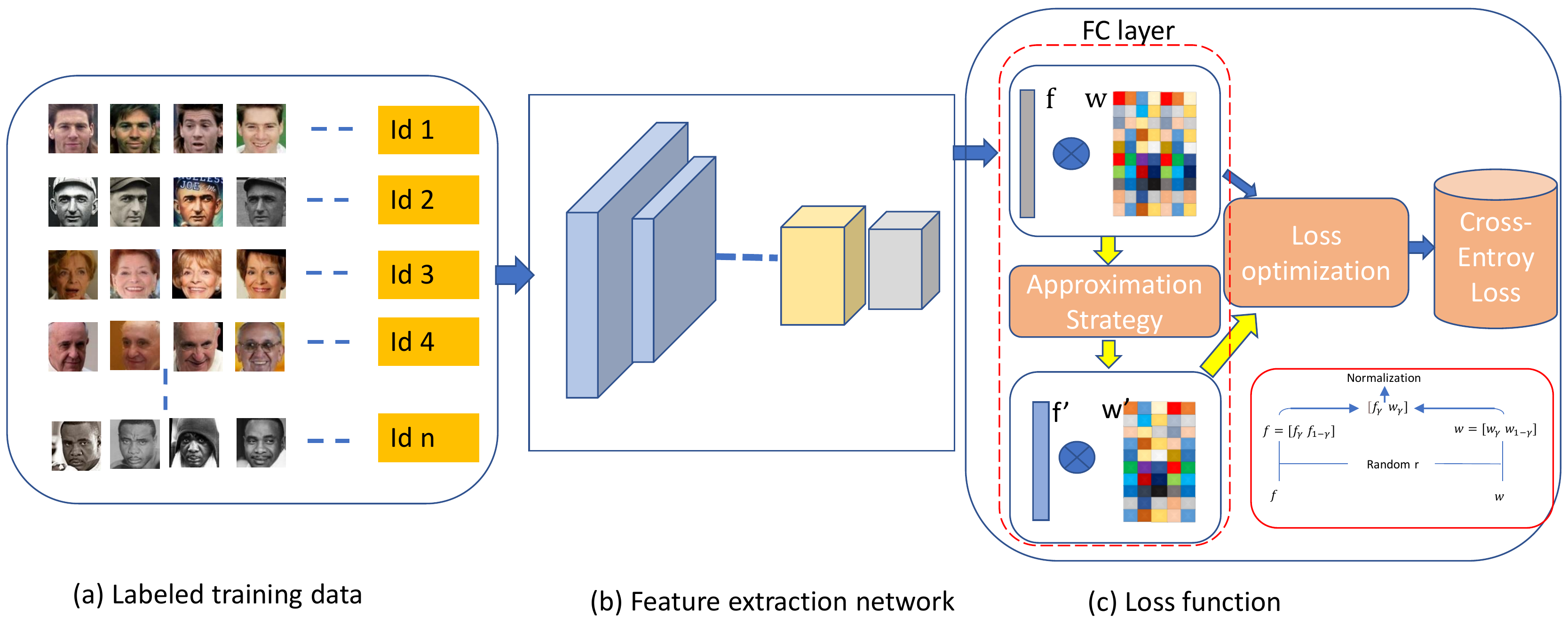}
  \caption{ The framework of the face recognition training system.  The main contribution of paper is highlighted by the red dashed box, and the step of approximation strategy is shown in red
  solid box. In the part of loss function, $f$ is the backbone embedding feature, $w$ is the linear transformation matrix in FC layer. The training path of most existing methods is
   carried out along the blue arrow. In this paper, we propose an approximation training strategy along the yellow arrow. $f^{'}$ and $w^{'}$ are used to replace $f$ and $w$ respectively.}
  \label{framework}
\end{figure}

\subsection{\textbf{Our Proposed Method}}
In this section, we present the approximation strategy to the face recognition training task. Firstly,
we illustrate the feature approximation mechanism in Section~\ref{section:A}, and then in Section~\ref{section:B}, we give more detailed information about the subfeature normalization strategy.

\subsubsection{\textbf{Feature approximation}}
\label{section:A}
In order to obtain the sampling features to realize softmax approximation, we propose a feature approximation strategy which can  realize accelerating or parallel
processing by sampling positive and negative samples, the framework of this paper is shown in Fig.~\ref{framework}.

Firstly, we present the details of our approximation method using the softmax loss, which can be formulated as follows:
\begin{equation}
  \label{eq1}
 L =-\frac{1}{N} \sum_{i = 1}^{N} log \frac{e^{w_{yi}^{T}f_{i}+b_{yi}}}{\sum_{j = 1}^{n}e^{w_{j}^{T}f_{i}+b_{j}}}
\end{equation}
where the inner product $ w_{j}^{T}f_{i}$ is considered as the distance between feature and weight.
As $w_j$ has the same dimension as $f_{i}$, which can be taken as a face representation, and meanwhile faces are locally distinguishable,
we use the inner product of subspaces to replace that in Eq.~\ref{eq1}.

On the basis of softmax loss, we introduce a dimension selection factor $r\in [0,1]$, the subfeature can be expressed as $R_t*f_{i}$, where $R_t \in {\left\{0,1\right\}^d}$ is a random matrix with $R_{(t,j)} \sim Bernoulli(r)$, $*$ denotes element-wise product. And $R_t*w_{yi}$ constitutes a subset of $w_{yi}$. Consequently, we get the approximate loss:
\begin{align}
\label{eq4}
\widehat{L} = -\frac{1}{N} \sum_{i = 1}^{N} log \widehat{p_{i}}
\end{align}
where $\widehat{p_{i}} = \frac{e^{(R_t*w_{yi})^{T}(f_{i})}}{\sum_{j = 1}^{n}e^{({R_t*w_{j}})^{T}(f_{i})}} $.

And when performing evaluation on downstream tasks, we still use the full feature as face representation. Through analyzing our approximation loss in Eq.~\ref{eq4},
obviously, $(R_t*w_{yi})^{T}f_{i} = w_{yi}^{T}(R_t*f_{i})$, so it can be easily expended as:

\begin{equation}
\label{eq5}
\widehat{L} = L + T_{Bernoulli}(r)
\end{equation}
$T_{Bernoulli}(r)$ is a term that depends on the sample ratio r, it is found that our approximation strategy has obvious regularization characteristics.

Compared with the original softmax loss function, our approximation strategy can dynamically replace the inner product of feature and weight by that of the subspace features.
When training with this approximation mechanism, the optimization goal for positive samples is to maximize the multiplication between the subfeature. Considering the randomness of this feature sampling, it focuses on making the aggregation of local features distribute evenly in the whole feature, while the original training method has no such effect.
For negative samples, there is no difference between global features and subspace features for enhancing inter-class distance.

\subsubsection{\textbf{Subfeature Normalization}}
\label{section:B}
According to the latest methods, the normalization of weight and feature has become the standard configuration. The normalization step makes CNN focus more on the optimization angle, and the obtained deep face features are more separated.
In practice, the bias term b in Eq.~\ref{eq1} is often set to 0, when the individual weight w and the feature $f_i$ are normalized.
The softmax loss function can be formulated as:
\begin{equation}
\label{eq2}
l_{softmax}=-\frac{1}{N} \sum_{i = 1}^{N} log \frac{e^{s\cos\theta_{yi}}}{s\cos\theta_{yi} + \sum_{j = 1,j\neq yi}^{n}e^{s\cos\theta_{j}}}
\end{equation}
$\theta_{yi}$ is the angle between the embedding feature $f_i$ and the center $w_{yi}$. The normalization makes the predictions of Eq.~\ref{eq2} only depends on the angle between feature and weight.
Given the angle classification boundary $\theta$, the convergence constraint of training process should meet:
\begin{equation}
\begin{cases} arcos(w_j^T f_i )<\theta  & if\  i=j \\  arcos(w_j^T f_i )>\theta & if \ i\neq j \end{cases}
\label{eq7}
\end{equation}


For the same purpose, we also hope that the optimization goal of this feature approximation training method will focus on the angle difference.
Considering with this,
we normalize the subfeature $R_t*f_i$ and $R_t*w_j$, and re-scaled $R_t*f_i$ to s,
the angle representation in Eq.~\ref{eq2} is replaced with the angle of between subfeature and subweight.
The angle constraint between $R_t*fi$ and $R_t*w_j$ should also meet:
\begin{equation}
\begin{cases} arcos(\omega _j^T x_i )<\theta  & if\  i=j \\  arcos(\omega _j^T x_i )>\theta & if \ i\neq j \end{cases}
\label{eq8}
\end{equation}
where $x_i=\frac{R_t*fi}{\parallel R_t*fi\parallel }$, $\omega_j=\frac{R_t*w_j}{\parallel R_t*w_j \parallel }$.



 \begin{algorithm}[!t]
   \caption{The proposed SubFace training method}
   \label{alg1}
   \begin{algorithmic}[1]
     \Require
       scale s, margin parameters $m_{1},m_{2},m_{3}$, class ids label, sample ratio r, feature f, class center w
     \Ensure
      Class-wise affinity score logits
     \State  ind= randperm(r)
     \State  x = indexselect(f,ind)
     \State  $\omega$ = indexselect(w,ind)
     \State  cosine = x*$\omega$
     \State cosine = norm(x)*norm($\omega$)
     \State  onehot = scatter(label,depth,on = 1.0,off = 0)
     \State  phi = $cos(arccos(m_{1}cosine) + m_{2})-m_{3}$
     \State  logits = s*(onehot*phi + (1-onehot)* cosine)
   \end{algorithmic}
 \end{algorithm}

 In this paper, we propose a new softmax approximation strategy, which utilizes the local separability of facial features. And the training procedure of our SubFace method is illustrated in the Alg.~\ref{alg1}.
 When adopting this approximation strategy to the margin-based loss, our method inherits the advantages of margin-based loss,
 and meanwhile, the discriminative ability of any subspace of embedded features is enhanced. Consequently, the co-adaptive relationship between feature is reduced.
 To the best of our knowledge, it is the first attempt to explore the softmax approximation to improve the performance of face recognition. Comprehensive experiments demonstrate that our method can achieve competitive performance compared with existing methods. It is worth mentioning that
our method only needs several lines of codes, which make it  more flexible and adaptable in real-world scenarios. 

 \section{Experiments}
 \label{sec4}

 \subsection{Datasets}
 \label{sec4.1}
We employ three widely used datasets CASIA~\citep{24}, A MS1M-RetinaFace~\citep{54}, MS1MV2~\citep{11} as our training datasets. The MS1MV2 contains 5.8M images and 85K identities. The MS1M-RetinaFace dataset contains 5.1M images of 93 K identities, which is a refined dataset of MS1M~\citep{53}.
And the CASIA dataset contains 10,577 identities and 0.5 M images. Note that all face images are resized to 112 $\times$ 112.

\subsection{Implementation Details}
We employ the widely used CNN architecture  MobilefaceNet~\citep{13}, ResNet-50, ResNet-100~\citep{1,2}  as the backbone and set the embedding feature dimmension to 512.
And during testing, two margin-based loss functions (ArcFace and CosFace) are used to make detailed comparison. We set the scale s to 64, the angular margin m to 0.5 when using ArcFace and the cosine margin m to 0.4 of CosFace. The mini-batch size is set to 512.
The SGD momentum is set to 0.9 and the weight decay is empirically set to 5e-4. The learning rate starts from 0.1 and is divided by 10 at 36k, 52k iterations for CASIA, and the total training process is finished at 65k iterations.
For MS1MV2, the learning ratio is divided at 100k, 160k iterations, the training process is finished at 180k iterations. For MS1M-RetinaFace, we divided the learning rate at 100k, 160k, 210k iterations and finished at 250k iterations. All experiments are conducted on a server equipped with two Nvidia A100 GPUs on Pytorch~\citep {paszke2019pytorch} framework.

\subsection{Important Parameter}
\begin{figure}[t]
  \centering
  \begin{minipage}[t]{0.33\linewidth}
  \centering
  \includegraphics[width=1.70in]{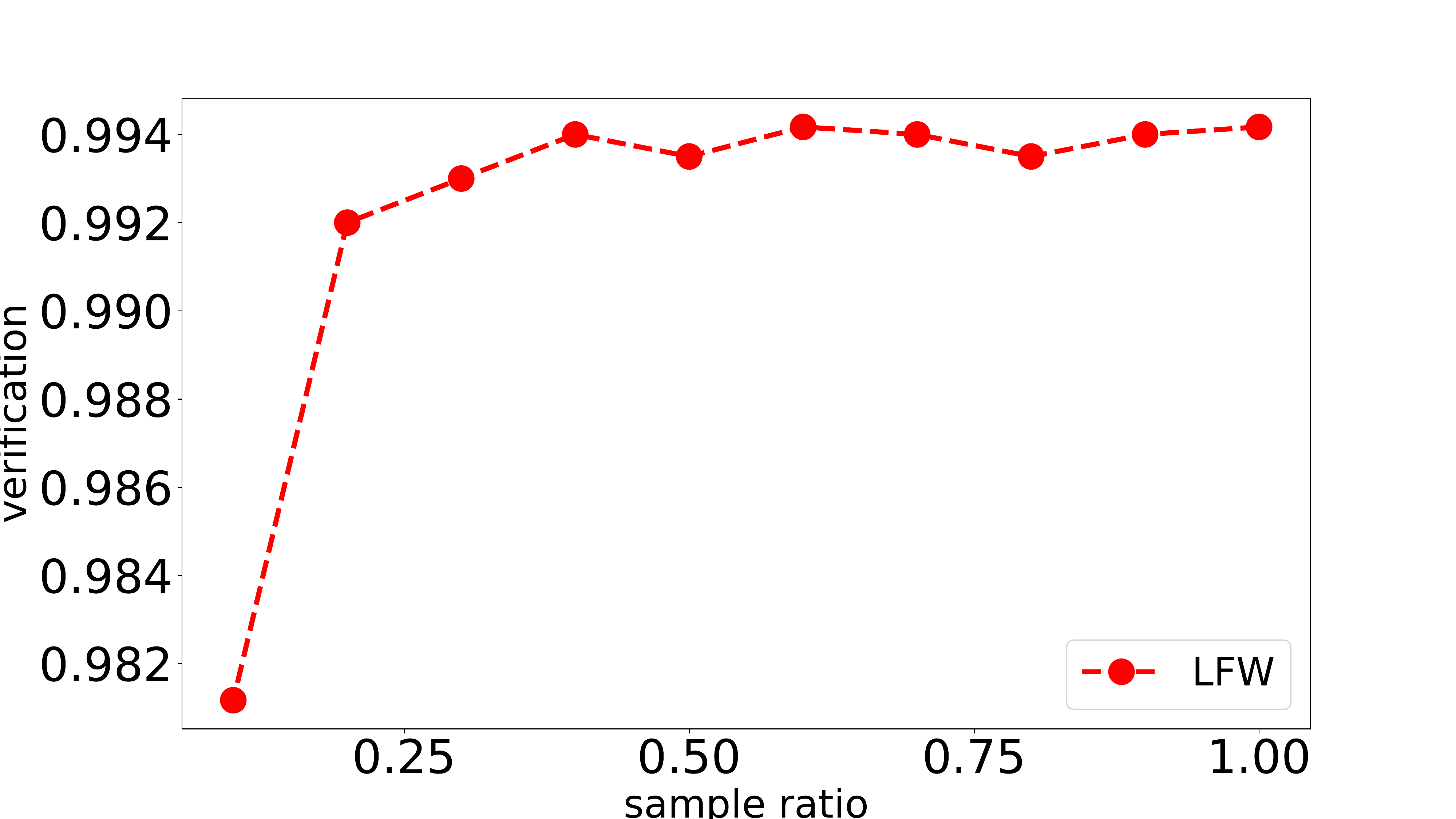}
  \centerline{(a)}
  \label{fig41}
  \end{minipage}%
  \begin{minipage}[t]{0.33\linewidth}
  \centering
  \includegraphics[width=1.70in]{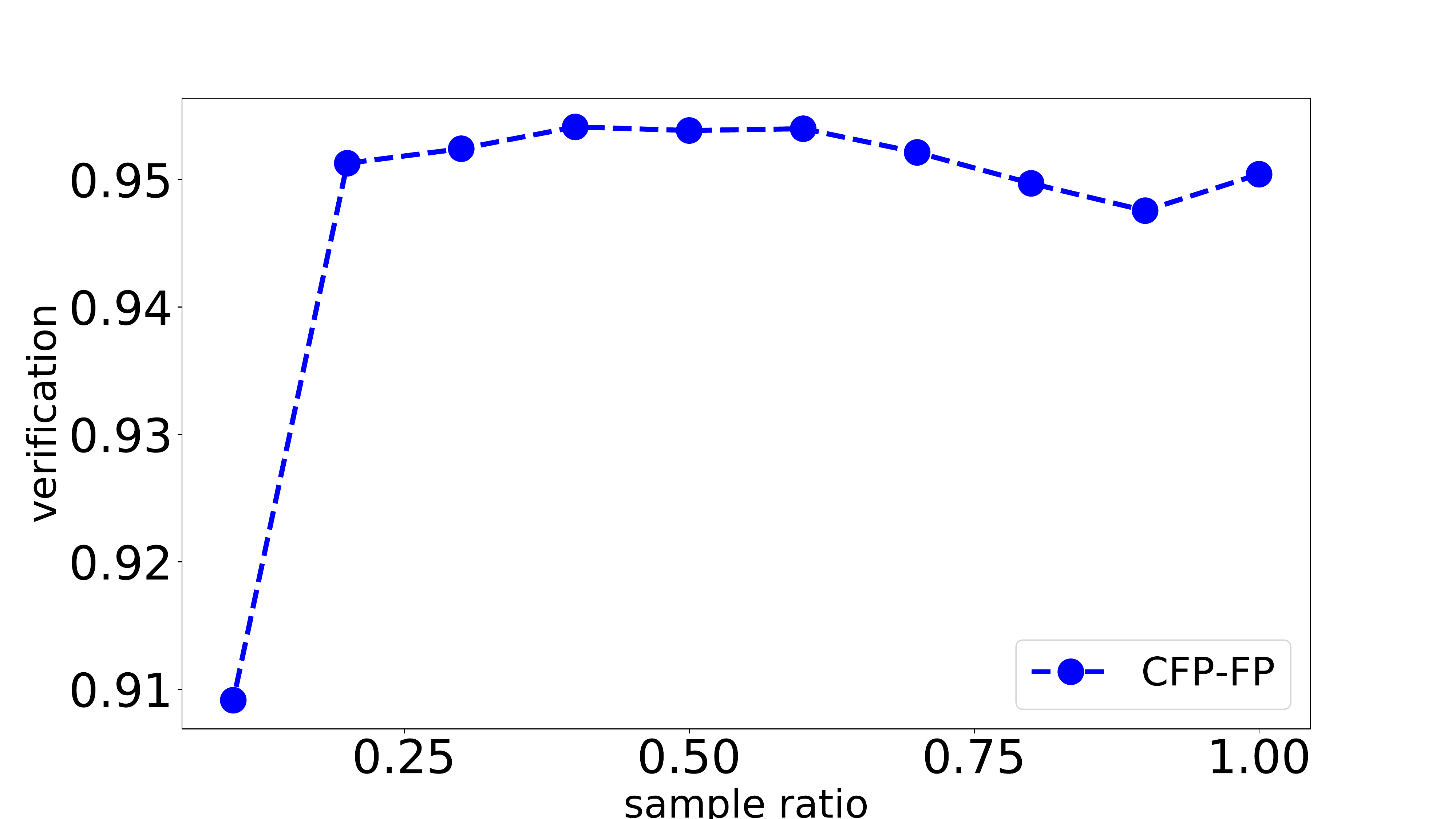}
  \centerline{(b)}
  \label{fig42}
  \end{minipage}
  \begin{minipage}[t]{0.33\linewidth}
    \centering
    \includegraphics[width=1.70in]{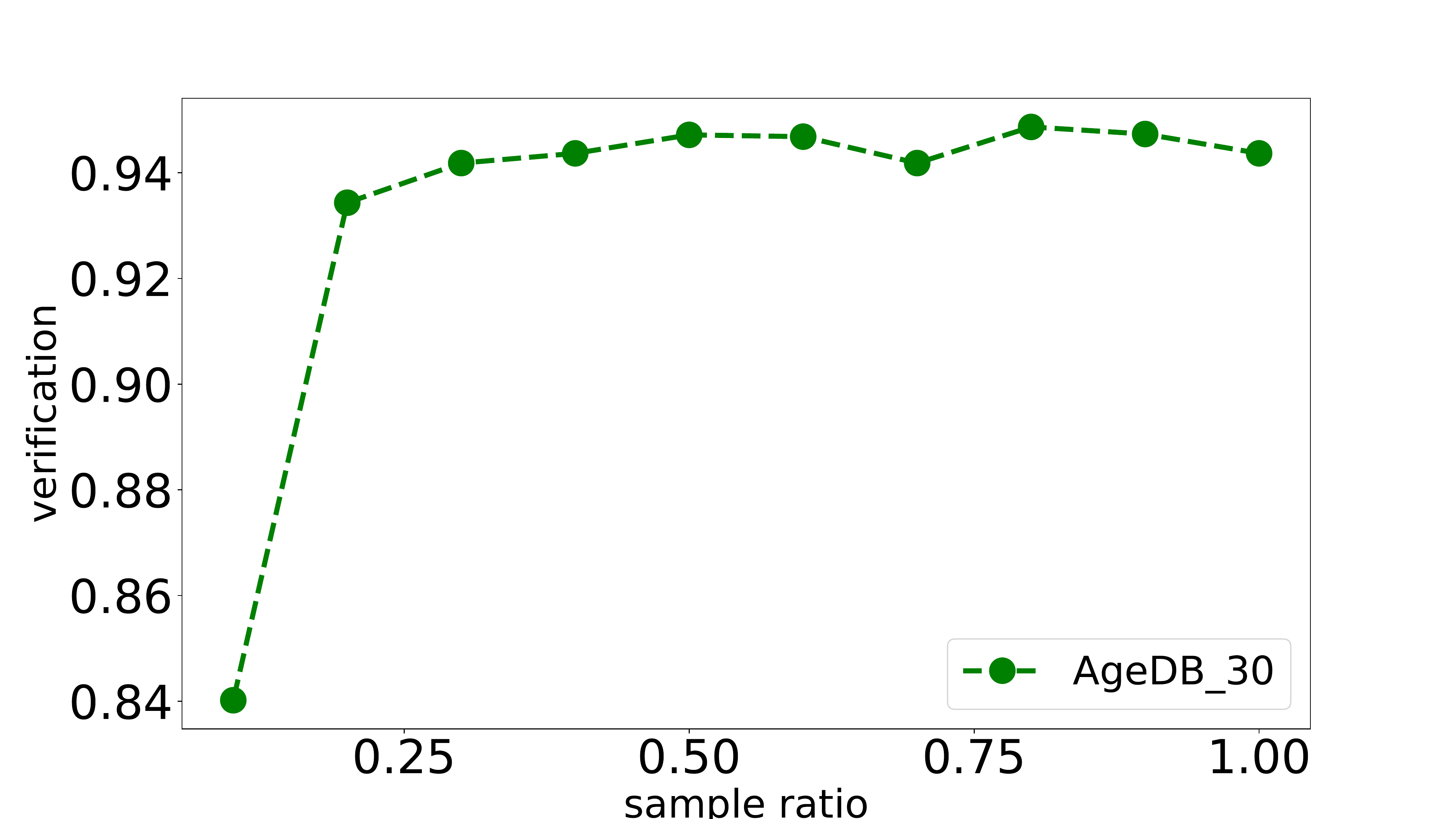}
    \centerline{(c)}
    \label{fig43}
    \end{minipage}
   \caption{Verification results on val datasets using different sampling ratio in the range of [0-1].  (a) Verification result on LFW. (b) Verification result on CFP-FP. (c)
   Verification result on AgeDB-30.}
  \label{fig4}
  \end{figure}

 We employ the ResNet-50 as the embedding network and the CASIA as training dataset to train ArcFace using our strategy,
 and we compare the verification result under different sampling ratio in the range of [0-1],
 as shown in Fig.~\ref{fig4}. The experiment shows that the performances of all
 sample ratio do no change a lot except when sampling ratio is low.
 For example, when the sampling rate is 0.1, it leads to a decrease in the recognition accuracy.
 More because the representation dimension of the feature is obviously insufficient. In this experiment,
 it means that only 51-dimensional features are used to represent the face during
 the training. And when the sampling ratio is greater than 0.4,
 our approach has little negative impacts on model performance,
 the performances on the three validation sets are more just fluctuating. And in order to make a fair comparison,
 we empirically set the sampling ratio at 0.7 in the follow experiments.

 \subsection{Ablation Study}
 \label{section:C}
To further validate the effectiveness of the our proposed method, we perform
several ablation studies on the individual component of our proposed SubFace method.

\begin{figure}[t]
    \centering
    \begin{minipage}[t]{0.33\linewidth}
    \centering
    \includegraphics[width=1.70in]{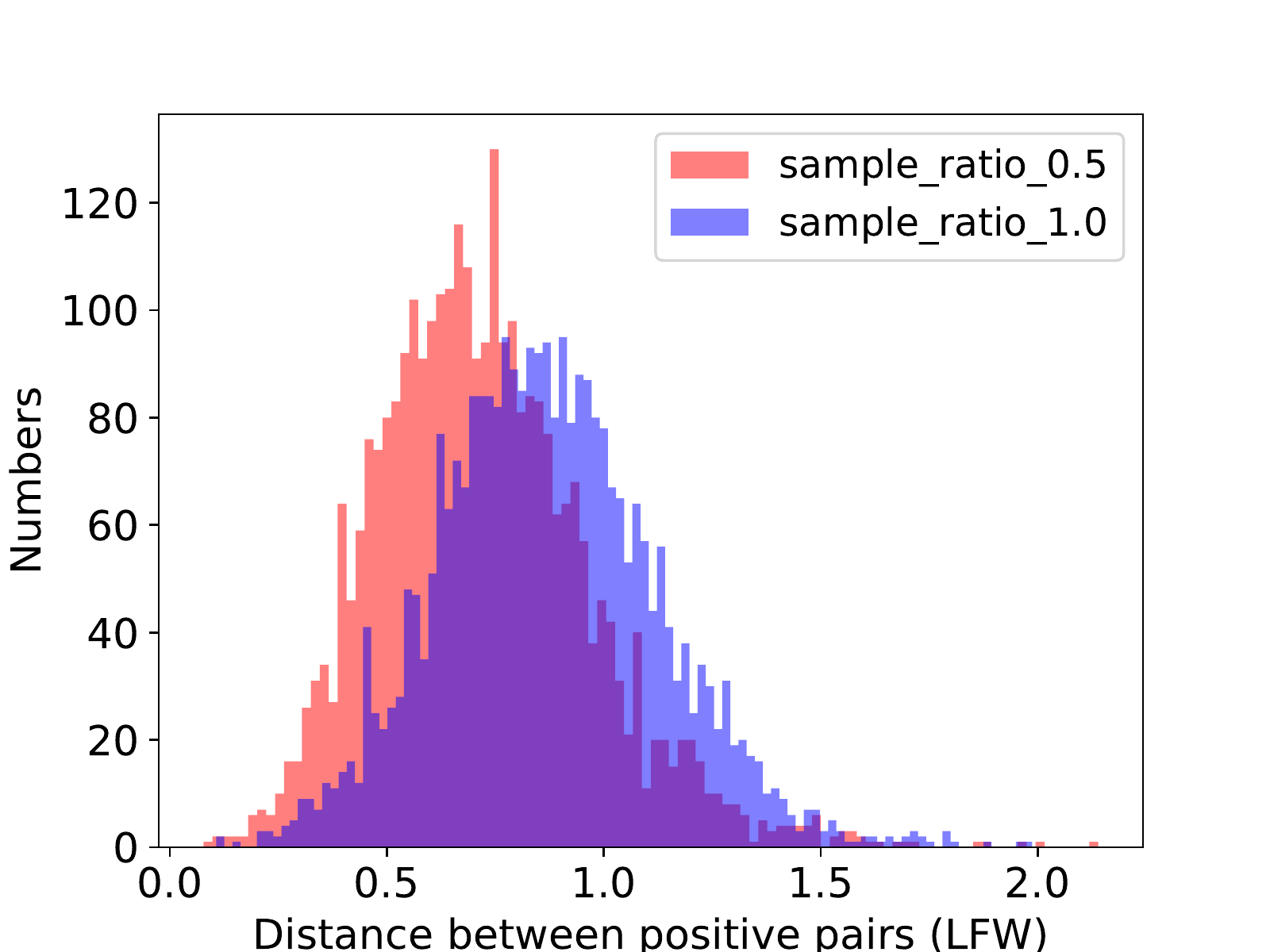}
    \centerline{(a)}
    \label{fig31}
    \end{minipage}%
    \begin{minipage}[t]{0.33\linewidth}
    \centering
    \includegraphics[width=1.70in]{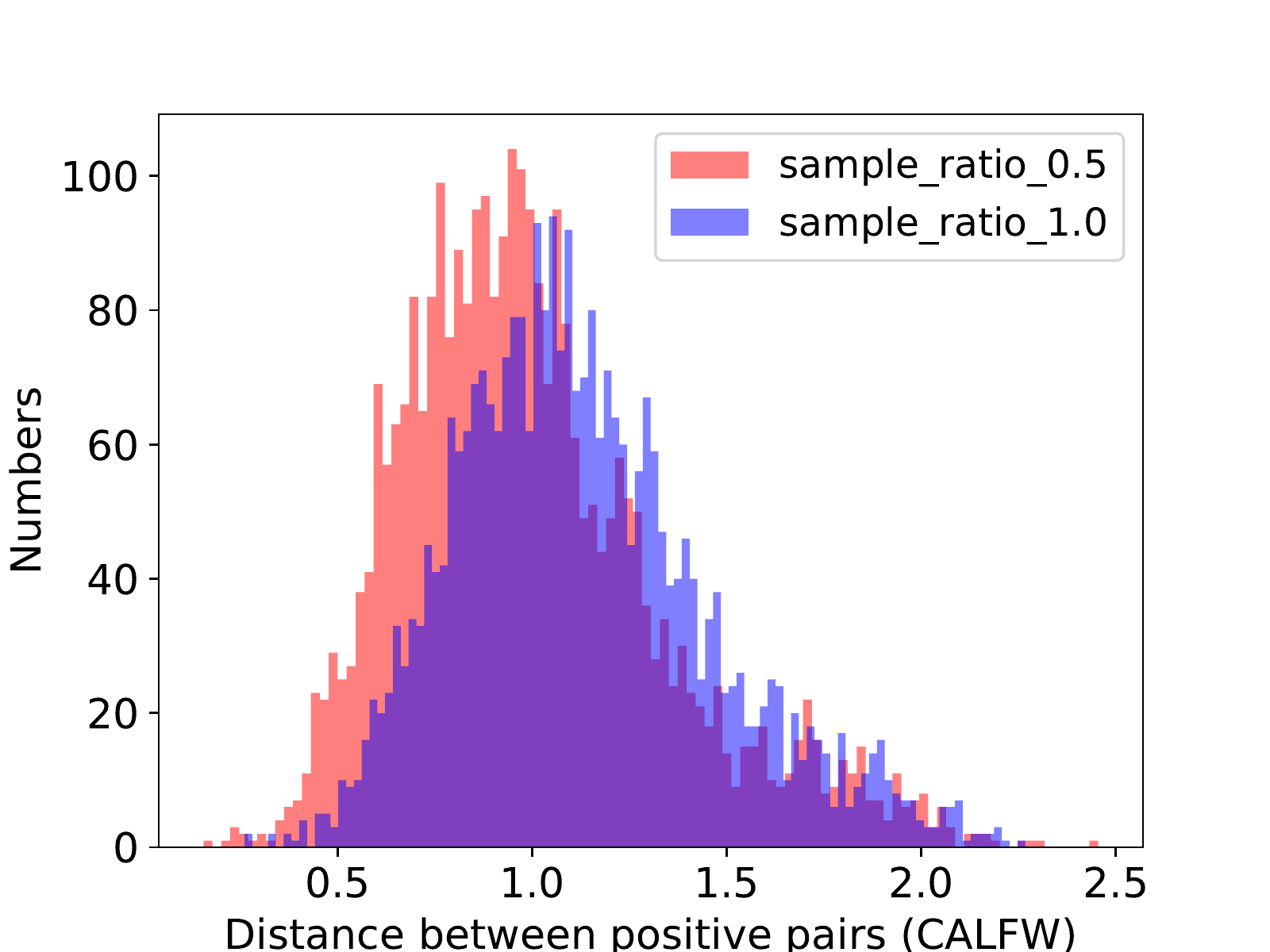}
    \centerline{(b)}
    \label{fig32}
    \end{minipage}
    \begin{minipage}[t]{0.33\linewidth}
      \centering
      \includegraphics[width=1.70in]{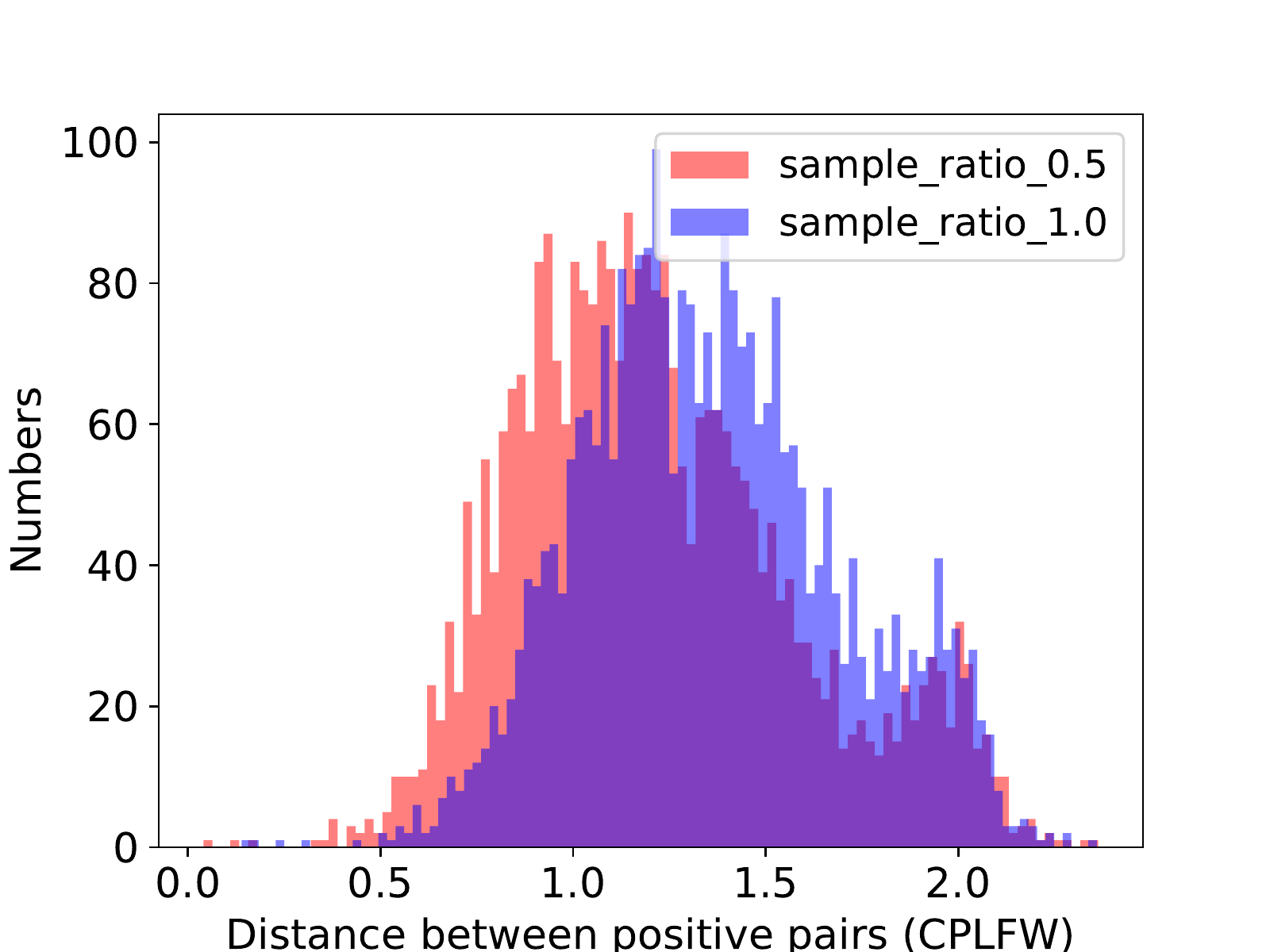}
      \centerline{(c)}
      \label{fig33}
    \end{minipage}
    \begin{minipage}[t]{0.33\linewidth}
      \centering
      \includegraphics[width=1.70in]{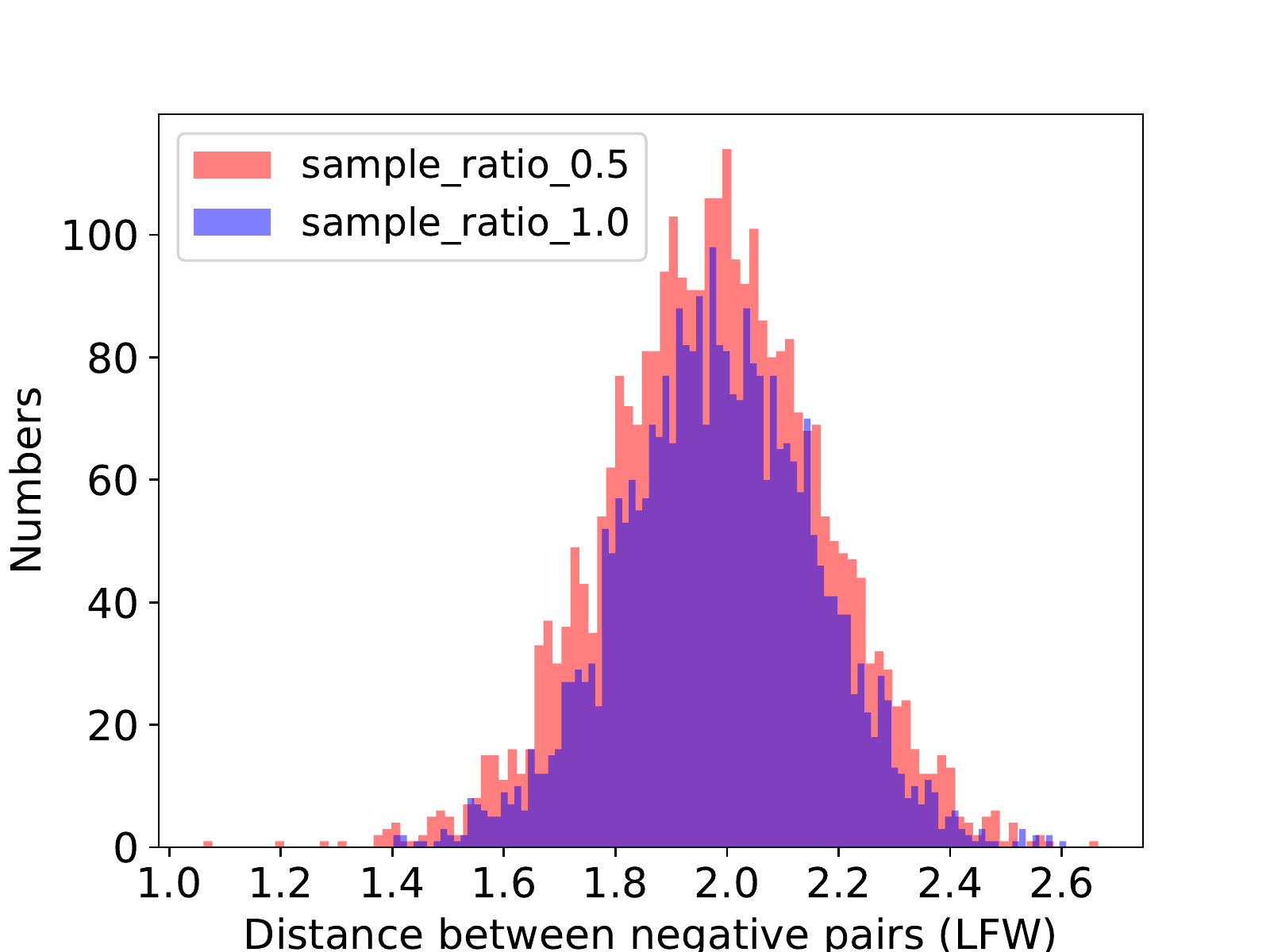}
      \centerline{(d)}
      \label{fig34}
      \end{minipage}%
      \begin{minipage}[t]{0.33\linewidth}
      \centering
      \includegraphics[width=1.70in]{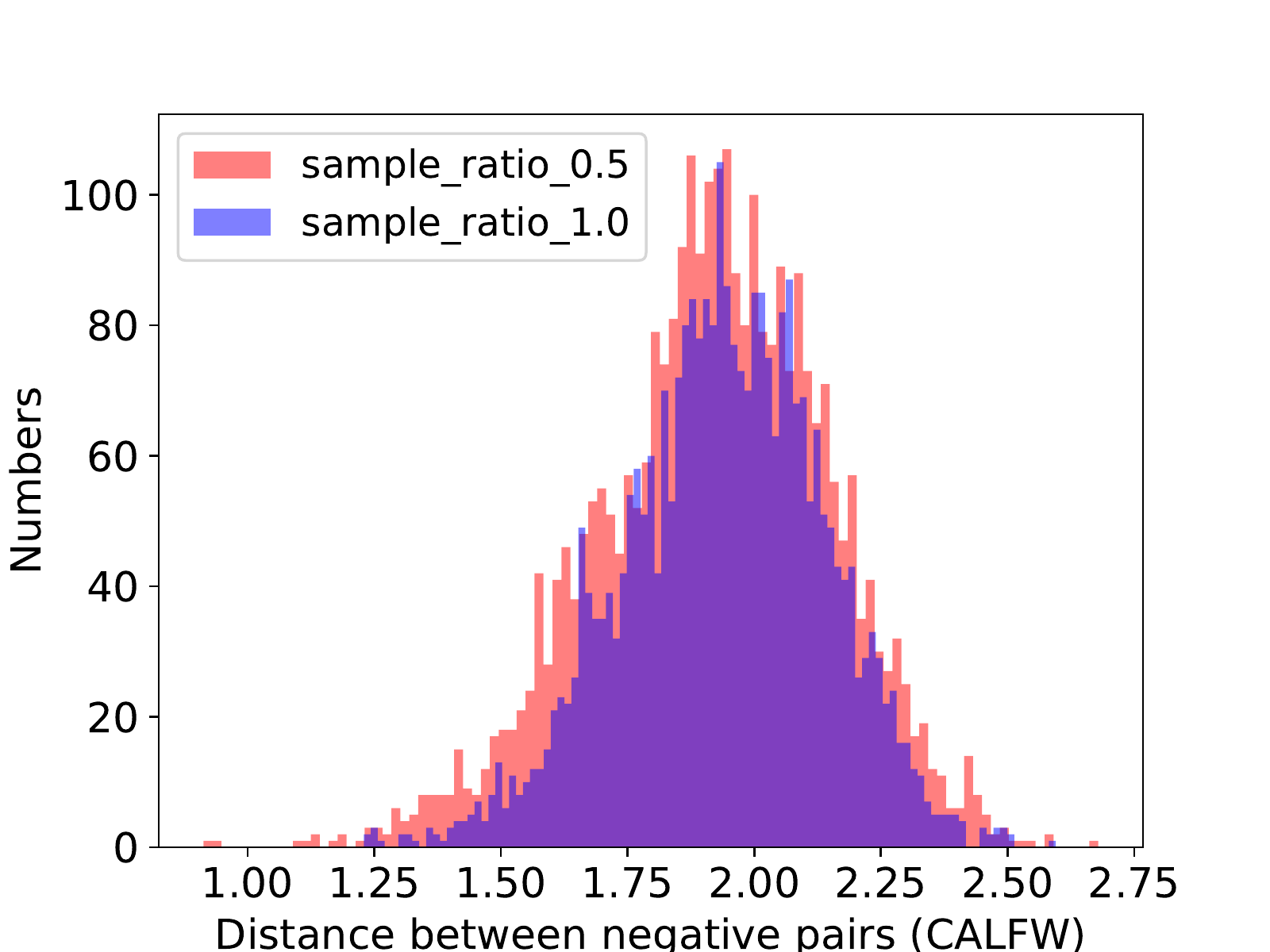}
      \centerline{(e)}
      \label{fig35}
      \end{minipage}
      \begin{minipage}[t]{0.33\linewidth}
        \centering
        \includegraphics[width=1.70in]{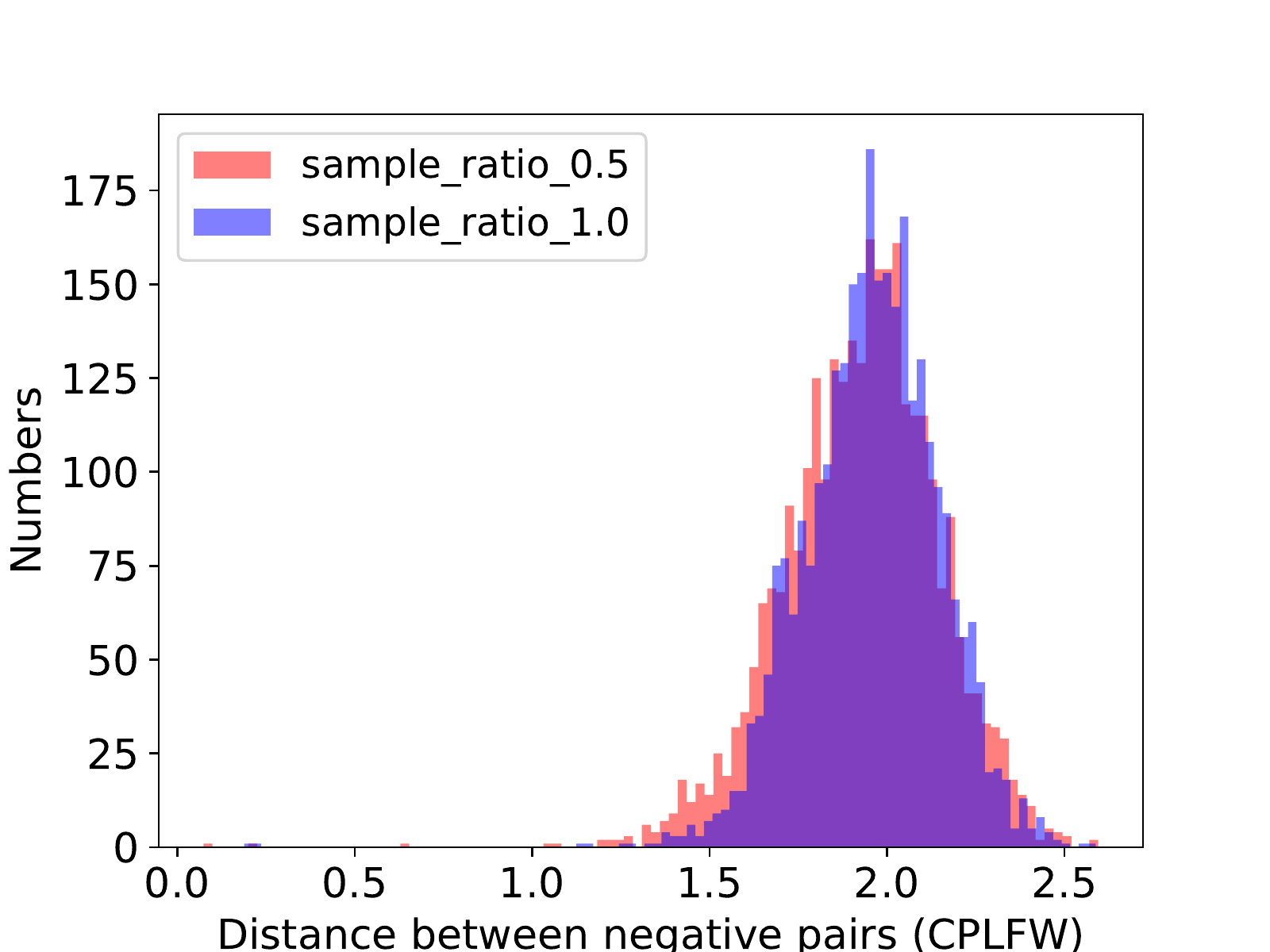}
        \centerline{(f)}
        \label{fig36}
      \end{minipage}
   \caption{The  distance distribution of face pair on LFW, CALFW, CPLFW dataset respectively.}
 \label{fig3}
\end{figure}

\textbf{Effects of feature approximation:} In this experiment, we test our approximation strategy on the validation dataset LFW, CALFW, CPLFW respectively with backbone ResNet-50, and we choose the ArcFace as the loss function. As illustrated in Fig.~\ref{fig3},  we give the feature distance distribution of face pairs. The distance distribution maps of face pairs at feature sampling rate 0.5 and 1.0  are given\footnote{The feature sampling rate of 1.0 represents the original training strategy.}. The Euclidean distances between positive pairs
are skewed to the smaller side compared with the situation at sampling ratio of 1.0. And there is no difference  of the negative sample pair under the two cases in terms of the distance distribution. Our random strategy makes the embedding features more compact, as the discriminative power of features is more balanced across
the entire dimension.

\textbf{Effects of subfeature normalization:} In this experiments we verify the convergence of our approximation strategy by observing the volatility of the vector angle.
 We denote $\theta_{j}$ as the angle between the embedding feature $f_i$ and the center $w_j$, $\theta_{j}^{'}$ as the angle between the subfeature $f_{(i,r)}$ and the subcenter $w_{(j,r)}$.
 In this experiment, we adopt the average cosine distance defined in~\cite{46} and judge the convergence by the variation of cosine.
 \begin{figure}[!t]
  \centering
  \includegraphics[width=8cm]{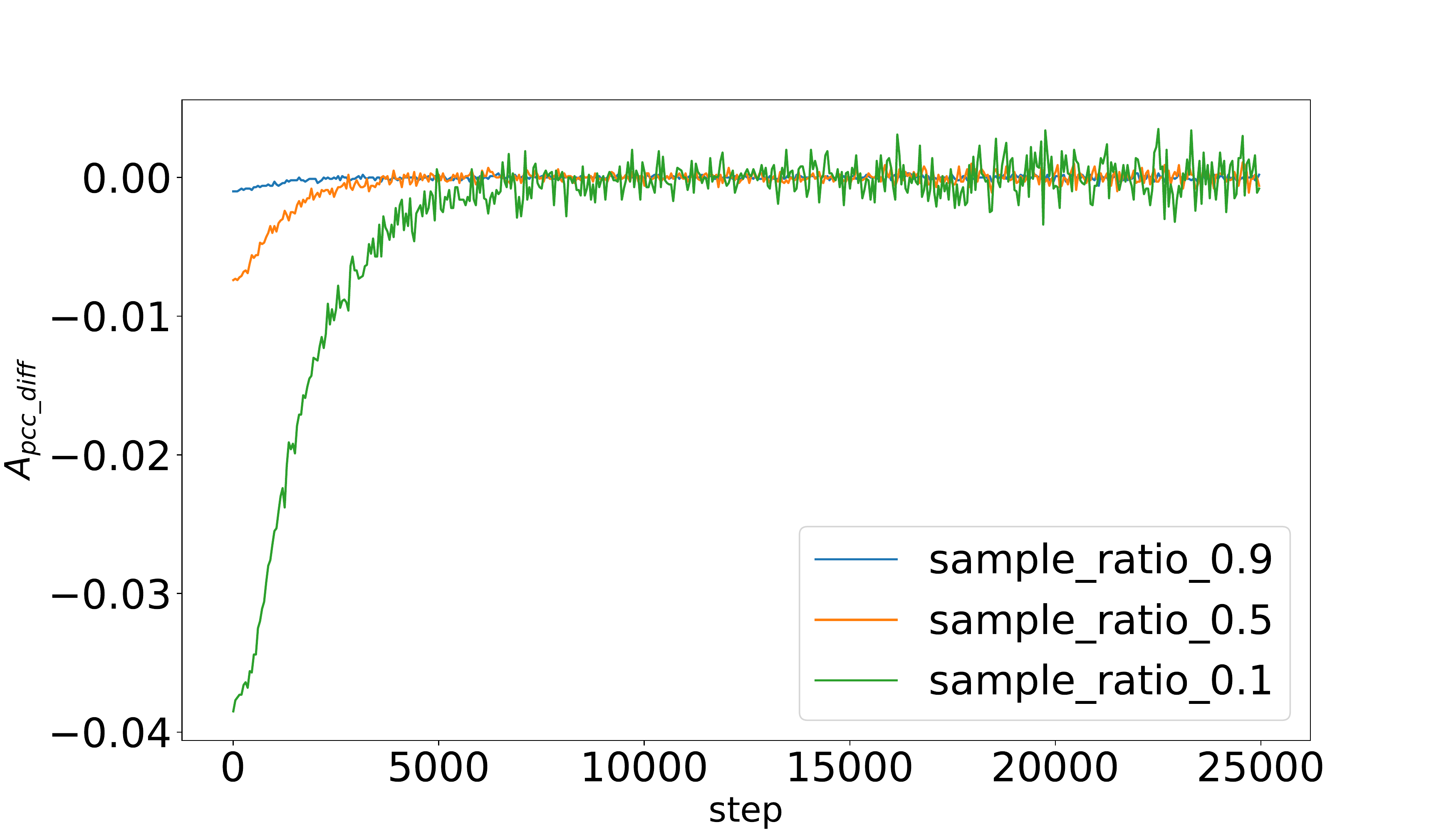}
  \caption{The average cosine distance when the sample ratio is 0.1, 0.5, 0.9 respectively.}
  \label{fig2}
\end{figure}
As illustrated in Fig.~\ref{fig2}, the change of sampling ratio does not affect the final convergence, and the arbitrary $r*d$ dimensional
subspace can approximate the performance of full-dimensional features. During the same time, we also observe that the change of sampling rate can affect the volatility of the average cosine distance. To be more specific, when the sampling rate is smaller, the deviation is larger at the initial stage of training and the volatility of diffs is greater at later stage.

\textbf{Effects of lightweight model:} Lightweight models, which are benefited from smaller network structures, fewer model parameters,
 as well as superior performance,
 are now widely used in the application of embedded devices, computer vision, \textit{etc.}
 In this section, we choose MobilefaceNet as the backbone network since MobilefaceNet is more balanced between
 computation, parameters and model performance.
 The parameter amount of Mobilefacenet is 0.99M, and the calculation amount is 439.8M (FLOPs),
 and it performs well on relevant data.
 We train ArcFace on CAISA and MS1M-RetinaFace respectively. According to the results in Table~\ref{Tab1},
 our approximation strategy shows the performance superiority on lightweight models. For example,
 it can achieve the best accuracy of 99.58\% on LFW  and 96.22\%  on Age-DB30 respectively when trained on MS1M-RetinaFace dataset.
 And when using CASIA dataset, our trained model also shows a competitive performance, outperforming the previous work~\citep{7} and surpassing the original result by 0.08\% on AgeDB-30 dataset.

 \begin{table}
   \centering
   \caption{Verification results on LFW, AgeDB-30 of different lightweight models}
   \footnotesize
   \setlength{\tabcolsep}{0.8mm}{
   \begin{tabular}{l c c c c}
     \toprule
     \multirow{2}{*}{Method} & Params & \multirow{2}{*}{LFW} & \multirow{2}{*}{AgeDB-30} & Training\\
     &(M)& & &Data\\
     \midrule
     MobileNetV1~\citep{49} &3.2& 0.9863&0.8895 &0.5M\\
     ShuffleNet~\citep{49}  &0.8& 0.9870&0.8927 & 0.5M\\
     MobilefaceNet~\citep{7} &0.99& 0.9928&0.9305 & 0.5M\\
     MobilefaceNet (CASIA, r=0.7) &0.99& 0.9928&0.9313 & 0.5M\\
     \midrule
     ShuffleFaceNet (0.5X)~\citep{47} &0.5& 0.9907&0.9245 & 5.1M\\
     MobilefaceNet~\citep{7}  &0.99& 0.9955&0.9607 & 3.8M\\
     MobilefaceNetV1~\citep{49}  &3.4& 0.9940&\textbf{0.9640} & 5.1M\\
     ProxylessFaceNAS~\citep{49} &3.2& 0.9920&0.9440 & 5.1M\\
     MobilefaceNet (MS1M-RetinaFace, r=0.7) &0.99& \textbf{0.9958}&0.9622 & 5.1M\\
     \bottomrule
   \end{tabular}}
  \label{Tab1}
 \end{table}

\textbf{Effects of small-scale \& large-scale trainset:} To validate the effects of small-scale \& large-scale trainset, we adopt the two training set CASIA and MS1MV2 to test the performance of our
 sampling method. We treat the CASIA as small-scaled trainset and the MS1MV2 as large-scaled trainset.
 In order to make a fair comparison and reduce the volatility of the test results themselves, we choose ResNet-100 as the backbone, and adopt the arcface loss and cosface loss for optimization process.

\begin{table}
   \centering
   \caption{Verification performance (\%) of ResNet-100 on LFW, CFP-FP, AgeDB-30 dataset respectively.}
   \footnotesize
   \setlength{\tabcolsep}{2.3mm}{
   \begin{tabular}{l c c c }
   \toprule
     Method & LFW  & CFP-FP & AgeDB-30 \\
     \midrule
     CASIA, ResNet-50, ArcFace(0.5)~\citep{11} &0.9953 & 0.9556&0.9515 \\
     CASIA, ResNet-50, CosFace(0.35)~\citep{11} &0.9951 & 0.9544&0.9456 \\
     \midrule
     MS1MV2, ResNet-100, ArcFace(0.5)~\citep{46} &0.9983 & 0.9845& 0.9820 \\
     MS1MV2, ResNet-100, CosFace(0.4)~\citep{46} &0.9983 & \textbf{0.9851}& 0.9803 \\
     \midrule
     CASIA, ResNet-100, ArcFace(0.5), SubFace (r=0.7) & {0.9952}  & {0.9584} & {0.9540}\\
     CASIA, ResNet-100, CosFace(0.4), SubFace (r=0.7) & {0.9950} & {0.9534} & {0.9485}\\

     \midrule
     MS1MV2, ResNet-100, ArcFace(0.5), SubFace (r=0.7)&{\textbf{0.9983}} &{0.9850}& {\textbf{0.9823}} \\
     MS1MV2, ResNet-100, CosFace(0.4), SubFace (r=0.7)&{0.9982}&{0.9837} &{0.9820} \\
     \bottomrule
   \end{tabular}}
  \label{Tab2}
 \end{table}

 According to the experimental results in Table~\ref{Tab2}, there are obvious performance gaps that the results on the big model (ResNet-100) are generally better than the results of the small model (ResNet-50). The results on the big-scale dataset are better than those on the small-scale dataset using the same model. And ArcFace generally outperforms CosFace under the same conditions using our strategy. For example, when using ResNet-100, ArcFace with our approximation strategy can achieve the best verification performance on AgeDB-30 dataset (\textit{e.g.} 98.23\%), surpassing the results~\citep{46} by 0.03\% with ArcFace(0.5)~\citep{46} when trained on MS1MV2.
 However, the performance difference between our SubFace strategy and ArcFace~\citep{46} is very slight on LFW and CFP-FP dataset respectively.

 \subsection{Comparison with the State-of-the-art Methods}
 In this section, we evaluate the model performance with different settings on the face verification datasets  such as LFW~\citep{26}, CFP-FP and agedb-30~\citep{28}.
 During the same time, we also give the performance of our methods on large-pose and large-age datasets CPLFW~\citep{29}, CALFW~\citep{30}, and the large-scale image datasets megaface~\citep{31}, IJB-B~\citep{23}, and IJB-C~\citep{35} respectively.

 \textbf{Evaluation on LFW, CALFW and CPLFW datasets:}
 LFW dataset is one of the widely used benchmarks for unconstrained face verification on images and videos,
 it contains 6,000 comparison pairs, with 3,000 positive pairs and 3,000 negative pairs. CPLFW and CALFW datasets are recently
 introduced which show higher pose and age variations with same identities from LFW.  According to the results from the Table~\ref{Tab3},
 ArcFace with our strategy
can  achieve the best performance of 99.85\% and 93.48\% on LFW and CPLFW respectively. When compared with the original ArcFace~\citep{46}, we have improved performance on all the three sets. Even with the results (the row 5 in Table~\ref{Tab3}) on a larger training sets Glint360k~\citep{46}, the performance is surpassed by +0.02\% and +0.09\% on LFW and CALFW datasets respectively.


 \begin{table}
 \centering
   \caption{Verification performance (\%) of different face recognition models on LFW, CALFW and CPLFW respectively.}
   \footnotesize
   \setlength{\tabcolsep}{1.5mm}{
   \begin{tabular}{l c c c }
   \toprule
     Method & LFW  & CALFW & CPLFW \\
     \midrule
     VGGFace2~\citep{50} & 0.9943 & 0.9057 & 0.8400 \\
     GroupFace~\citep{51} & 0.9985 & 0.9620 & 0.9317 \\
     CurricularFace~\citep{52} & 0.9980 & 0.9620& 0.9313 \\
     MS1MV2, ResNet-100, ArcFace~\citep{46} &0.9982 & 0.9545 & 0.9208  \\
     Glint360k, ResNet-100, CosFace~\citep{46} &0.9983 &0.9621 & \textbf{0.9478} \\
     \midrule
     MS1M-RetinaFace, ResNet-100, ArcFace, SubFace (r = 0.7) &{\textbf{0.9985}} &{ \textbf{0.9630}} & {0.9348}  \\
   \bottomrule
   \end{tabular}}
  \label{Tab3}
 \end{table}

 \textbf{Evaluation on IJB-B and IJB-C datasets:}
 The IJB-B dataset contains 1,845 subjects with 21.8 k still images and 55 k frames from 7,011 videos.
 The IJB-C contains 3,531 subjects with 31.3 k still images and 117.5 k video frames.
 And there are 10,270 genuine matches and 8m imposter matches in the IJB-B verification protocol.
 The IJB-C verification protocol provides a total 19,557 genuine matchs and 15,639K impostor matches.
 \begin{table}
 \centering
   \caption{ The 1:1 verification accuracy(TAR@FAR=1e-4) on the IJB-B and IJB-C dataset respectively.}
   \footnotesize
   \setlength{\tabcolsep}{4.2mm}{
   \begin{tabular}{l c c}
   \toprule
     Method & IJB-B  &  IJB-C  \\
     \midrule
     MS1MV2, ResNet-100, ArcFace~\citep{11} &0.942 & 0.956\\
     MS1MV2, ArcFace~\citep{46}&0.948 & 0.962\\
     GroupFace~\citep{51} & 0.949 & 0.963 \\
     CurricularFace~\citep{52} & 0.948 & 0.961 \\
     \midrule
     MS1MV2, ArcFace, ResNet-100 (SubFace, r=0.7)&{0.9501 }& {0.9638}\\
     MS1M-RetinaFace, ArcFace, ResNet-100 (SubFace, r=0.7)&{\textbf{0.9547}} & {\textbf{0.9685}}\\
     \bottomrule
   \end{tabular}}
  \label{Tab4}
 \end{table}
 On the IJB-B and IJB-C datasets, we employ the MS1M-retinaFace and MS1MV2 dataset as the training data and the ResNet-100 as the embedding
 network, the sampling ratio is choosen at 0.7 for the fair comparison with
 the most recent methods. As shown in Table~\ref{Tab4}, our method further improves the performace
 the TAR (@FAR=1e-4) to 96.85\% and 95.47\% on IJB-C and IJB-B respectively. And when using the same training dataset MS1MV2, the accuracy difference between
 our method and ArcFace~\citep{11} is 0.81\% (94.2\% vs 95.01\%) and 0.78\% (95.6\% vs 96.38\%) on IJB-B and IJB-C respectively, which surpasses the performance in CurricularFace~\citep{52} by a clear margin.



 \textbf{Evaluation on MegaFace dataset:} Finally, we evaluate the performance on the MegaFace Challenge.
 The evaluation protocol of MegaFace includes gallery and probe sets.
 The gallery set contains 1M images of 690k different individuals and the probe set contains 100 k photos of 530 unique individuals from facescrub~\citep{36}.
 As there are some noises in the original Megaface, we adopt the refined megaface dataset~\citep{11} to make a fair comparison.
 Table~\ref{Tab5} shows the results obtained by ResNet-100 and the previous works for both identification and verification tasks on this dataset.
 From the result we can see that ResNet-100 with our strategy can achieve state-of-the-art results with respect
 to other recent models under verification scenarios.

 \begin{table}
  \centering
  \caption{ Face identification and verification evaluation of different methods. These experiments are tested on Megafce Challenge 1 using FaceScrub as the probe set. ``id" refers to the rank-1 face identification accuracy with 1e6 distractors, and ``ver" denotes to the face verification TAR at 1e${-6}$ FAR. }
   \footnotesize
   \setlength{\tabcolsep}{3.4mm}{
   \begin{tabular}{l c c}
   \toprule
     Method & id  &  ver  \\
     \midrule
     ArcFace(0.5)~\citep{11} &0.9835 & 0.9848\\
     CosFace(0.35)~\citep{11} &0.9791 & 0.9791\\
     CASIA, ResNet-50, ArcFace~\citep{11}&0.9175 & 0.9369\\
     MS1MV2, CosFace~\citep{46} & 0.9836 & 0.9858\\
     MS1MV2, ArcFace~\citep{46} & 0.9831 & 0.9859\\
     CurricularFace~\citep{52} & \textbf{0.9871} & 0.9864 \\
     \midrule
     MS1M-RetinaFace, ResNet-100 ,ArcFace (SubFace, r=0.7)&{0.9839} & {\textbf{0.9871}}\\
     \bottomrule
   \end{tabular}}
  \label{Tab5}
 \end{table}

\subsection{Discussion}
In this work, we propose a face training method based on local feature approximation, which has achieved competitive results on several benchmark datasets. On the one hand, compared with existing training methods,
we mainly use the local feature for training, and we also verify the convergence and feature compactness of this method in Section~\ref{section:C}.
Unfortunately, the sampling dimension corresponding to the subfeature is consistent with all data in the same batch, which may not be an optimal choice in same cases.
For example, the distinctive local area of the face may be different between profile and frontal, or the face under the condition of illumination and non-illumination. Therefore, there is still room for further optimization in the selection of our random sampling method.

On the other hand, we perform in-depth empirical analysis of our method, and it is surprising to observe that the training optimization is difficult to reach a comprehensive optimization state in all performance indicators, which leads to a slightly performance degradation on some benchmark datasets. For example, as shown in Table~\ref{Tab2}, the verification performances of our method on LFW and CFP-FP datasets are 99.83\% and 98.50\% respectively.
While the best results of previous methods on LFW and CFP-FP datasets are 99.83\% and 98.51\% individually. The reasons for this phenomenon may come from the following aspects:

First, the sampling factor in our experiment is set to 0.7, maybe this optimal parameter should be different in various experiments.

Second, according to the Table~\ref{Tab5}, the performance gap between our method and existing methods in some datasets (MegaFace dataset) is relatively small, where the fluctuation of this difference can be completely compensated through code, training strategy and data enhancement, \textit{etc}.

Third, the main purpose of our optimization method is to promote the compactness of positive sample features, so as to improve the performance of face classification. In the comparative study in Table~\ref{Tab2}, the performance difference here may be mainly resulted from the difference of the processing method on some hard samples. And these challenges warrant further research and consideration when deploying the face recognition model in real scenarios.

\section{Conclusion}
 \label{sec5}

 In this paper, we propose an approximate training strategy named SubFace to enhance the distinguishing ability of features. It uses the subspace of the class center combined with the subfeature to achieve intra-class compactness.
 Comprehensive experiments conducted on benchmarks demonstrates that SubFace can significantly improve the performance of vanilla CNN baseline with margined-based loss on face recognition, proving its superiority and competitiveness when compared with the state-of-the-arts. For further research, we will combine other optimization strategies, such as distributed computing, and neural architecture search, to further improve the performance of the proposed method.

\bmhead{Acknowledgments}

This work was supported by the National Natural Science Foundation of China under Project (Grant No. 81974276).
The authors would like to thank the anonymous reviewers for their valuable suggestions and constructive criticisms.

\section*{Declarations}

\begin{itemize}
\item \textbf{Funding} \\  This work was supported by the National Natural Science Foundation of China under Project (Grant No. 81974276).
\item \textbf{Conflict of interest} \\  The authors declare that they have no conflict of interest.
\item \textbf{Ethics approval} \\  Not Applicable. The datasets and the work do not contain personal or sensitive information, no ethical issue is concerned.
\item \textbf{Consent to participate} \\  The authors are fine that the work is submitted and published by Machine Learning Journal. There is no human study in this work, so this aspect is not applicable.
\item \textbf{Consent for publication} \\  The authors are fine that the work (including all content, data and images) is published by Machine Learning Journal.
\item \textbf{Availability of data and material} \\  The data used for the experiments in this paper are available online, see Section~\ref{sec4.1} for more details.
\item \textbf{Code availability} \\  The code will be publicly available once the work is published upon agreement of different sides.
\item \textbf{Authors' contributions} \\  Hongwei Xu and Suncheng Xiang contributed conception and design of the study, as well as the experimental process and interpreted model results. Dahong Qian obtained funding for the project and provided clinical guidance. Hongwei Xu and Suncheng Xiang drafted the manuscript. All authors contributed to manuscript revision, read and approved the submitted version.
\end{itemize}


\bibliography{sn-bibliography}


\end{document}